\documentclass[sigconf]{acmart}

\usepackage{booktabs} 
\usepackage{todonotes} 
\usepackage{url}
\usepackage{comment}
\usepackage{subfig}

\setcopyright{rightsretained}

\copyrightyear{2018} 
\acmYear{2018} 
\setcopyright{acmlicensed}
\acmConference[FDG18]{Foundations of Digital Games 2018}{August 7--10, 2018}{Malm\"o, Sweden}
\acmBooktitle{Foundations of Digital Games 2018 (FDG18), August 7--10, 2018, Malm\"o, Sweden}
\acmPrice{15.00}
\acmDOI{10.1145/3235765.3235792}
\acmISBN{978-1-4503-6571-0/18/08}

\begin{document}
\title{DATA Agent}
\subtitle{}

\author{Michael Cerny Green}
\email{mcgreentn@gmail.com}
\affiliation{%
  \institution{Tandon School of Engineering, New York University}
  \city{New York City}
  \state{NY}
}

\author{Gabriella A. B. Barros}
\email{gabbbarros@gmail.com}
\affiliation{%
  \institution{Tandon School of Engineering, New York University}
  \city{New York City}
  \state{NY}
}

\author{Antonios Liapis}
\email{antonios.liapis@um.edu.mt}
\affiliation{%
  \institution{Institute of Digital Games, University of Malta}
  \city{Msida}
  \country{Malta}
}

\author{Julian Togelius}
\email{julian@togelius.com}
\affiliation{%
  \institution{Tandon School of Engineering, New York University}
  \city{New York City}
  \state{NY}
}


\begin{abstract}
This paper introduces DATA Agent, a system which creates murder mystery adventures from open data. In the game, the player takes on the role of a detective tasked with finding the culprit of a murder. All characters, places, and items in DATA Agent games are generated using open data as source content. The paper discusses the general game design and user interface of DATA Agent, and provides details on the generative algorithms which transform linked data into different game objects. Findings from a user study with 30 participants playing through two games of DATA Agent show that the game is easy and fun to play, and that the mysteries it generates are straightforward to solve. 
\end{abstract}

\maketitle

\section{Introduction}

Procedural content generation (PCG) in games is now a flourishing research field as well as a well-received game design practice~\cite{shaker2016procedural,short2017procedural}. Among the more ambitious visions in PCG is the automatic or semiautomatic design of complete games~\cite{cook2014rogue,togelius2008experiment}. Many ideas for how to create games are purely algorithmic, imagining that games can be created through computational procedures with no particular input, \emph{ex nihilo} if you wish. But human game designers do not create games out of nothing; such games are embedded in a cultural context and constantly inspired by their experiences, including other games, high-flying ideas and mundane goings-on.

Luckily, much of the real world now exists digitally represented in easily accessible form. It makes sense that freely available data could be used to inform content generation and game generation algorithms, in effect helping them transform the real world into games. Such data-driven game content generation has been attempted under the moniker ``data games'' \cite{friberger2012generating}. In particular, the \emph{Data Adventures} series
has explored ways of automatically creating adventure games in the style of \emph{Where in the world is Carmen Sandiego?} (Br{\o}derbund Software, 1985) from data sources such as Wikipedia and OpenStreetMap (OSM)~\cite{haklay2008openstreetmap}.

Previous iterations of the Data Adventures series produced games that successfully integrated data from various sources, but were somewhat lacking in playability. This was deemed to be partly due to the user interface, but also due to the core game design patterns that the generator used. In order to convincingly demonstrate the potential for data-driven adventure game generation, the generated games need to be enjoyable and intuitive, necessitating a redesign of both the game interface and the game generator.

This paper presents the outcome of that process, DATA Agent, the third generation installment from the Data Adventures series. DATA Agent has a redesigned backstory, trying to create coherent scenarios while acknowledging the frequently absurd results of building on Wikipedia data; a new story generation mechanism to work with the new story pattern; and a redesigned user interface. The games produced by the DATA Agent generator, while having in common with previous Data Adventures games that they are adventure games generated from open data, play very differently. Below, we detail the game design and generator design of DATA Agent, and also the results of a user study where 30 individuals played two games generated by DATA Agent and answered a questionnaire. Key findings from the user study include the fact that players were curious to play more mysteries with different people, the interface and general interactions were intuitive, and that it was straightoforward to find the solution to each mystery.


\section{Background}\label{sec:related}

The DATA Agent system generates playable adventure games based on open data, building on prior work in open data techniques and procedural content generation as well as the design conventions of adventure games. Below, we describe existing work on \emph{data games} in Section \ref{sec:related_datagames}, survey the definition and game design of adventures games in Section \ref{sec:related_adventure}, and review various projects involving story, quest, and dialog generation (as core components of adventure games) in Section \ref{sec:related_story}.

\subsection{Data Games}\label{sec:related_datagames}

\emph{Data games} use freely available information to automatically generate game content \cite{friberger2012generating}. This term was first proposed by \citeauthor{friberger2012generating}, describing a generator that transforms open data into \emph{Monopoly} (Hasbro, 1935) boards ~\cite{friberger2012monopoly}. Data games promote the visualization and interaction of information in creative and/or interesting ways. Bar Chart Ball, for example, uses demographic data selected by the player and represented as a bar chart~\cite{togelius2013bar}; the game is played by choosing what is visualized. 
Similarly, Open Trumps generates decks of \emph{Top Trumps}\footnote{\url{https://en.wikipedia.org/wiki/Top_Trumps}} cards using governmental information~\cite{cardona2014open}; and geographical data from OpenStreetMap (OSM)~\cite{haklay2008openstreetmap} have been used to generate maps and players' initial positions for FreeCiv\footnote{FreeCiv is an open source version of Civilization.  (\url{http://www.freeciv.org/})}~\cite{barros2015balanced}.

A different type of data game is proposed in \emph{A Rogue Dream} ~\cite{cook2014rogue}, which uses the auto-complete results of Google queries (using templates) to generate names for player abilities, enemies and healing items of a rogue-like game. The resulting game entities are not stemming from structured data in open-access repositories, but rather from the crowd-sourced data machine-learned by Google algorithms. Notably, sprites for A Rogue Dream entities are found via Google image searches using the word describing each entity. In MuseumVILLE\footnote{MuseumVILLE: (\url{https://github.com/bogusjourney/museumville/})}, content is selected from Europeana\footnote{Europeana is a portal for accessing digitised cultural heritage material, such as paintings and books, from more than 2,000 institutions across Europe.}. The player of MuseumVILLE takes the role of a museum curator who must theme their museum based on their interests. \emph{911 Operator} (Jutsu Games, 2016) is a commercial game about operating emergency telephone lines. The city maps used in the game are real-world maps, detailed up to exact street names and intersections.

Generating games from data allows for new play experiences to be shaped, for a multitude of purposes such as learning, contemporaneous or personalized games. Data-driven design opens the possibility for maximalist games that draw inspiration and combine content from dissimilar sources and points of view \cite{barros2018maximalist}.

\subsection{Adventure Games}\label{sec:related_adventure}
According to Fern\'andez-Vara's definition, an \emph{adventure game} is a simulation in which the player interacts with the rule system of a fictional world, populated with a series of concatenated puzzles which structure the performance of the player~\cite{vara2009tribulations}. Adventure games as a genre include a variety of game types, including \emph{interactive fiction}, \emph{graphical text adventures}, and \emph{point-and-click games}.

Mystery adventure games, a subtype of adventure games, cast the player into the role of a detective and gameplay revolves around solving a crime or mystery. In these games, players often visit exotic locations around the world and speak to a variety of interesting characters to gather clues and evidence. Games within this genre include \emph{Where in the World is Carmen Sandiego?} which tasks the player with finding a fugitive criminal, and the \emph{Sherlock Holmes} series (Frogwares 2002). Other games, such as \emph{Indiana Jones and the Fate of Atlantis} (LucasArts 1992), the \emph{Tomb Raider} series (Eidos 1996), and \emph{Uncharted} (Naughty Dog 2007), see the player embark on an adventure to solve mysteries and puzzles, often facing down criminal organizations and adversaries along the way.  

\subsection{Story, Quest, and Dialog Generation}\label{sec:related_story}
Research into story, quest, and dialog generation has mostly dealt with the creation and organization of text. Veale explored the value of using characters familiar to most audiences and auto-generating written stories with them~\cite{vealedeja}. \emph{M\'{E}XICA} is a system capable of generating creative writing~\cite{perez2001mexica}; it was further enhanced by combining an nn Narrator~\cite{montfort2007generating} to create \emph{M\'{E}XICA}-nn, a story generator capable of creating high-level plot lines~\cite{montfort2008integrating}. \emph{MABLE} is a generative ballad machine, combining the creation of stories and lyrics~\cite{singh2017ballad}, which is built on top of \emph{M\'{E}XICA}. \emph{Chronicle} is a system that generates unpredictable stories that are still coherent~\cite{pickering2017applying}. \emph{BRUTUS} creates dark stories of betrayal using characters with complex backgrounds and narratives~\cite{bringsjord2000artificial}. \emph{MINSTREL} uses author-level problem solving to generate stories about King Arthur and his knights~\cite{Turner:1993:MCM:166478}. 

In adventure games, quests and dialog are an essential part of the narrative. \emph{Mystery of Solaris} uses a two-tiered generator that constructs maps and missions together~\cite{lavenderadventures}. \emph{Charbitat} uses a lock-and-key mechanism to create a sense of progression in the game~\cite{alderman2006charbitat,ashmore2007quest}. \emph{Symon}, a point-and-click adventure game that uses PCG to create puzzles, was expanded into the ``Puzzle Dice System'' as a generalized way to generate puzzles in adventure games~\cite{fernandez2012procedural,fernandez2014creating}. 

Work on dialog generation can trace its roots back at least to \emph{ELIZA}, an early natural language processing computer which could simulate conversation~\cite{weizenbaum1966eliza}. \emph{Text2dialog} transforms monological text into dialog, and agents then act it out~\cite{hernault2008generating}. Interactive fiction uses dialog generation and delivery extensively, e.g. combining narrative goals, implicit forms of character expression, and emotional relationships between characters to generate realistic dialog \cite{cavazza2005dialogue}. 

Creating dialog, quests, and narrative for games is part of a larger subject of transforming an input from one medium to another. Probably the most fully realized game with interactive drama is Fa\c{c}ade \cite{mateas2003faccade}. The core interaction mode in Fa\c{c}ade is dialog, taking natural language inputs from the user and transforming it into Non-Playable Characters' (NPCs) reactions. In another example, Scheherazade~\cite{li2015scheherazade} uses crowd-sourcing to create generalized models of domains, such as bank robberies, for interactive storytelling.

\begin{figure*}[t]
\centering
\subfloat[Snapshot of the intro video]{\includegraphics[width=0.32\textwidth]{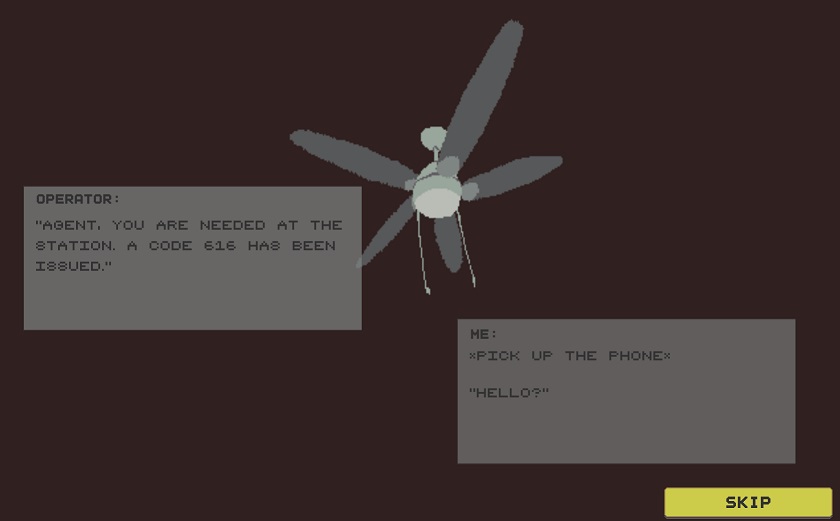}\label{fig:backstory_cutscene1}}\quad
\subfloat[Start Menu]{\includegraphics[width=0.32\textwidth]{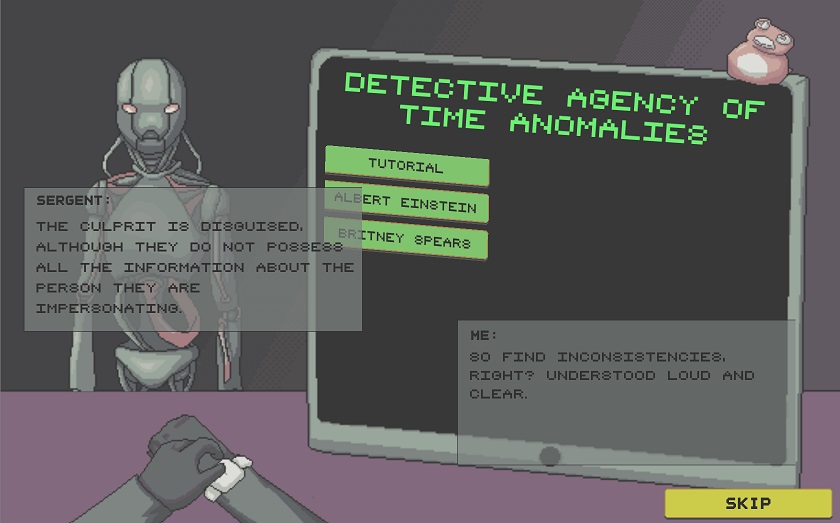}\label{fig:backstory_intro}}\quad
\subfloat[Game Lost Screen]{\includegraphics[width=0.32\textwidth]{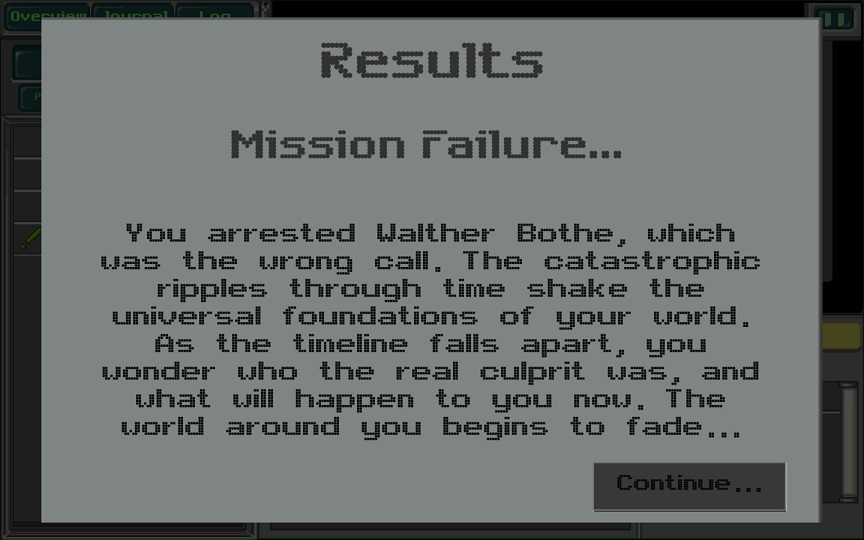}\label{fig:backstory_outro}}
\caption{An extensive introductory cutscene (Fig.~\ref{fig:backstory_cutscene1}) frames the time anomaly backstory and DATA agency. The dialog on the start menu screen (Fig.~\ref{fig:backstory_intro}) provides a short reminder on the context and winning conditions of the game. The winning or losing screens (Fig.~\ref{fig:backstory_outro}), at the end of the game, also tie in with the time anomaly backstory.}
\label{fig:backstory}
\end{figure*}


\section{The Data Adventure series}\label{sec:series}

DATA Agent is the third installment in the Data Adventures series. The goal of this series is to explore how semantically linked open data can be used to generate structured point-and-click adventure games \cite{barros2018maximalism}. Open data is exploited to not only generate individual game elements, such as NPCs and locations, but also to link them together as causal links that can be revealed through player agency. While different game generators within the series use different narratives and technologies, the underlying design principle is for a player to explore and unlock parts of the story as they visit and interact with game elements; the core technological principle is using DBPedia (a structured database derived from Wikipedia) for finding associations between people, using OpenStreetMap (OSM) for finding the maps for in-game locations and Wikimedia Commons for images of NPCs and items.

The first instance in the Data Adventures series is the generator of the same name \cite{barros2015data}. Data Adventures builds linear experiences, where players must find a target NPC starting from the house of another NPC. Both the initial and goal NPCs are created from real-world people with Wikipedia articles. A path between these NPCs is discovered through links from their semantically linked DBPedia entries. The linear path between them serves as input for the automatic creation of the game, from places to NPCs and dialog.

WikiMystery~\cite{barros2016murder,barros2018killed} is the successor to Data Adventures which can generate entire point-and-click murder mysteries using minimal human assistance. After being given a single name as input, the system can build a mystery game where a virtual representation of the given person was killed. The player has to travel to places, finding clues and talking to NPCs in order to find suspects and discover the culprit. Like Data Adventures, a core concept in WikiMystery is its link to real-world data accessed via open data repositories. Everything in WikiMystery corresponds to some snippet of open data, be it the physical locations of cities on the world map or the background of the building the player was currently located.


\section{DATA Agent}\label{sec:game}

DATA Agent is a system that combines a murder mystery game generator (as it generates games from open data) and a game: in the game, the player must identify a time-traveling doppelganger who has committed a murder and is now masquerading as a historical person. In the game, the player must travel between places and interact with historical people (NPCs), in order to collect information which will allow them to identify the culprit in a lineup of suspects based on historical people. The premise of a murder mystery is intended to prompt the curiosity of the player; having more than one possible suspects which the player must learn facts about also increases the game's learning potential.
The architecture of the DATA Agent system is divided into client-side and generator-side. The client side (i.e. the actual game) was built using the \emph{Unity 3D} game engine (Unity Technologies, 2005), while the generator-side (i.e. the generator of adventures) was built in Java. The generator creates adventures using an evolved version of past Data Adventures generators. The client-side game design is detailed in Section \ref{sec:design}; the server-side generator is summarized in Section \ref{sec:generator}.

\subsection{Game Design}\label{sec:design}

The following sections describe the narrative and visual design of DATA Agent, as well as a writeup of how the game is played.

\subsubsection{Backstory}\label{sec:design_backstory}

DATA Agent introduces a crafted narrative to explain the context and links between entities in the game. This serves both to explain certain expected inconsistencies (e.g. people from different time periods existing in the same game) and to clarify that both the murderer and the culprit do not refer to real events or people, respectively. The inconsistencies arise from the messiness of open data, and this will be discussed further in Section \ref{sec:discussion}. The player takes the role of a detective working for the Detective Agency of Time Anomalies (DATA). In the game's universe, criminals can go back in time and murder famous people, altering the time line and creating inconsistencies. To prevent this, DATA sends agents to the past to catch the killer and prevent the murder. Since the very act of going back in time messes up time lines, DATA has incomplete information about the past, and can only tell the player who the suspects are. It is the job of the player to learn facts about the suspects from other NPCs and objects, then catch the culprit, who is attempting to impersonate one of the suspects. Thankfully, the culprit has incomplete information about the person they are impersonating, and therefore will lie to the player about who they are. If the player has all the correct information about that individual, they can catch the culprit red-handed and save the day. All of this is clarified in the introductory cut-scene (which can be skipped), during the introductory dialog before the player starts the game, and at the end of the game: see Fig.~\ref{fig:backstory} for clarifications on how this narrative is communicated.

\begin{figure*}[t]
\centering
\subfloat[NPC Dialog]{\includegraphics[width=0.32\textwidth]{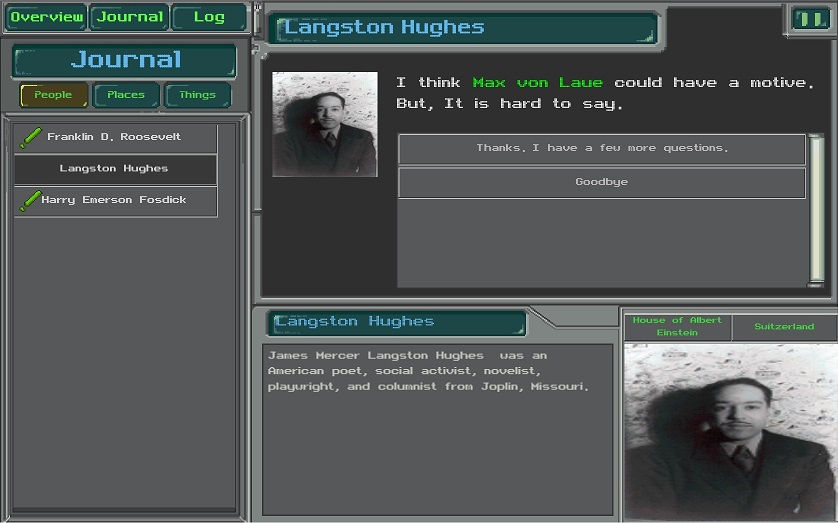}\label{fig:d1}}\quad
\subfloat[City screen, where buildings can be selected.]{\includegraphics[width=0.32\textwidth]{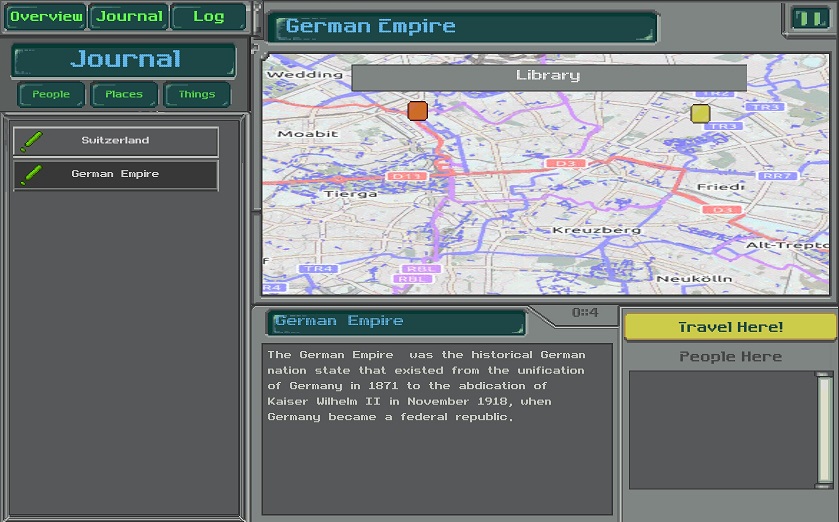}\label{fig:d2}}\quad
\subfloat[Item interaction]{\includegraphics[width=0.32\textwidth]{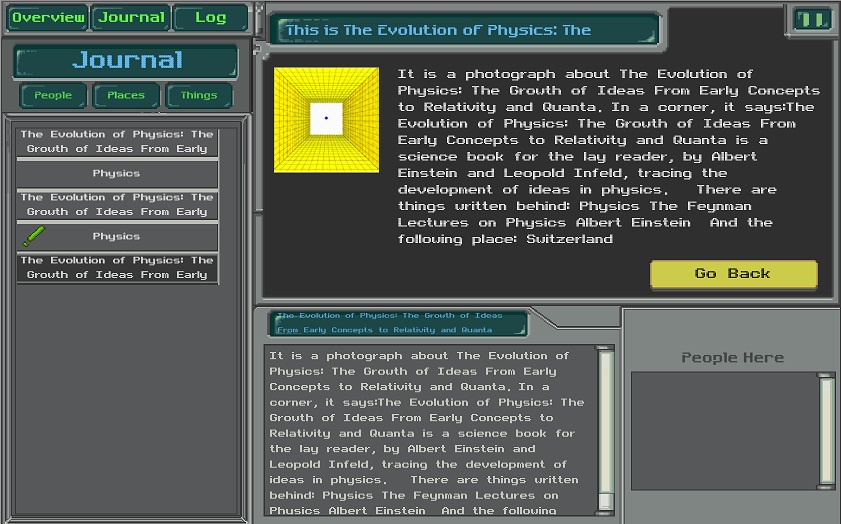}\label{fig:d3}}\quad
\\
\subfloat[Collectible key]{\includegraphics[width=0.32\textwidth]{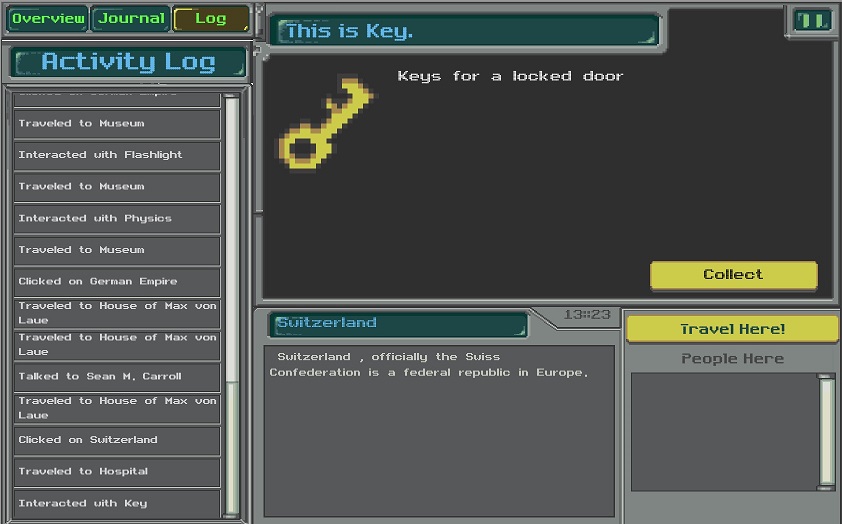}\label{fig:d4}}\quad
\subfloat[Building that can be unlocked with the key]{\includegraphics[width=0.32\textwidth]{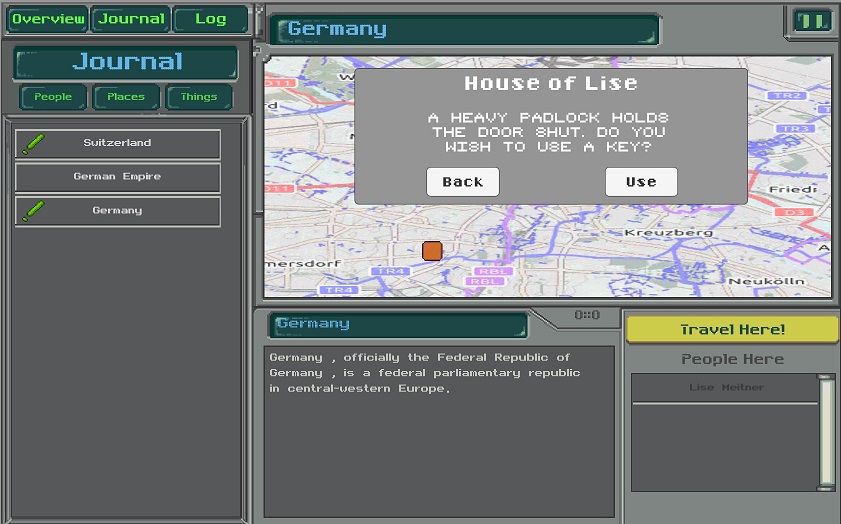}\label{fig:d5}}\quad
\subfloat[Interrogation screen]{\includegraphics[width=0.32\textwidth]{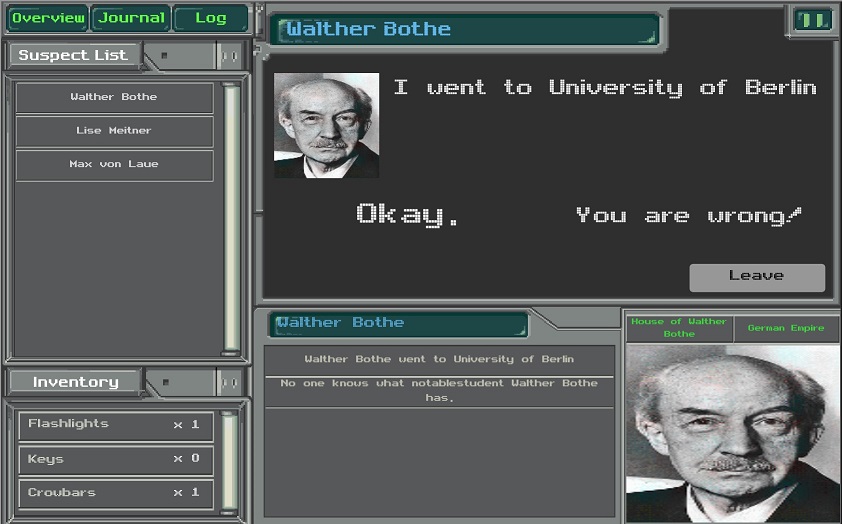}\label{fig:d6}}
\caption{Different screenshots of the DATA Agent User Interface, from the playtsted Albert Einstein game.}
\label{fig:interface}
\end{figure*}

\subsubsection{User Interface}\label{sec:design_interface}
The Unity client, which allows players to interact with the game, was designed to allow for ease-of-use. The goal is for players to be able to access at any point all  information they have gained throughout the game. Fig. \ref{fig:interface} shows several screenshots of the game. The game interface is split into three different displays: the game display (top right) for most game interactions, the description panel (bottom right) for showing information relevant to the current interaction, and the side-panel (left) which has several tabs displaying all information gathered so far. 

\begin{itemize}
\item The \emph{Game Display} shows the player's current location and represents the current activity, e.g. talking to a person (see Fig.~\ref{fig:d1}), looking at a city map (see Fig.~\ref{fig:d2}), or reading a book (see Fig.~\ref{fig:d3}). 
\item The \emph{Description Panel} contains a description of the last person, place, or object the player clicked on, either accessed from the Game Display or from a tab of the side-panel. If the player clicked on a place, they can travel to this place by clicking the ``Travel To'' button (see Fig.~\ref{fig:d2}). Places also display a list of known individuals the player found at this location. If the player clicked on a suspect from the Overview Panel, a list of known facts is displayed instead (see Fig.~\ref{fig:d6}). 
\item The \emph{Journal Tab} marks all people (see Fig.~\ref{fig:d1}), places (see Fig.~\ref{fig:d2}), and objects (see Fig.~\ref{fig:d3}) that the player has uncovered, distinguishing those that have not been fully explored.
\item The \emph{Activity Tab} displays in-game actions the player has taken, including traveling to locations, talking to people, and inspecting items for clues. This allows the player to remember if they have already performed some actions and/or exhausted the possible actions in a location. The side-panel of Fig.~\ref{fig:d4} shows the activity tab.
\item The \emph{Overview Tab} displays a list of suspects and current items in the player's inventory (top and bottom of the side-panel in Fig.~\ref{fig:d3} respectively). If the player clicks on a suspect, the Description Panel displays all known information about this suspect as well as their known whereabouts (see Fig.~\ref{fig:d6}).
\end{itemize}

As noted above, DATA Agent provides the player with a journal containing notes about people, places and objects that they encounter in the game, as shown in Fig. ~\ref{fig:d1}. When a player encounters something for the first time, the journal provides a glowing notification and an exclamation mark near the newly added journal entry for the player's convenience. Clicking on an entry will display information in the Description Panel. This was designed to act as an easily-referenced encyclopedia in case the user wants more information about a specific game item. Any fact discovered about a suspect is also written to the journal for the player to easily access during an \textit{interrogation}, described in the Section \ref{sec:design_gameloop}. The addition of the journal and the activity log helps a user keep track of their past actions and discoveries, as well as the tasks that they still have to do (i.e. new journal entries); this addresses usability concerns in past games in the Data Adventures series.

\subsubsection{Game Loop}\label{sec:design_gameloop}

Players must talk to people and inspect items to learn clues and facts about suspects in order to win the game. In this version of DATA Agent, dialog reveals \emph{facts} about suspects, and both dialog and items reveal \emph{clues} for new cities and buildings to explore. Along the way, the player may encounter buildings they are unable to enter unless they spend a consumable item to unlock it. For instance, the player may come across a locked building (see Fig.~\ref{fig:d5}) that needs a key. Only after finding a key (see Fig.~\ref{fig:d4}) and using it at the locked location will the player be able to enter and inspect the items and people within.


After finding a suspect (initially their locations are hidden from the player), they may talk to him/her to start an \emph{interrogation}. During an interrogation, the suspects list facts about themselves, to which the player can respond ``Okay'' or ``You are Wrong!'' (see Fig.~\ref{fig:d6} and \ref{fig:s3}). Facts mentioned by the suspect are the same ones that the player should have acquired from other NPCs in the game. If a player presses ``Okay'', the suspect continues to list another fact; if the player presses ``You  are Wrong!'' then a pop-up message allows the player to ``Arrest This Person.'' The culprit of the murder always lists one incorrect fact, i.e. a fact that differs from what other NPCs have provided to the player as facts for this suspect. Specifically, during interrogation the culprit lists a fact that is true for another suspect rather than a true fact about themselves. For example, if the culprit is Lise Meitner, when interrogated she will state that she does not have a notable student (which is true for the other suspect Walther Bothe); if the player has collected the true fact about Lise Meitner ``notableStudent: Max Delbr\"{u}ck'', they can identify that she is the time-traveling doppelganger. If the player chooses to arrest the suspect, the game ends. If this suspect was the culprit, a message appears congratulating the player, otherwise another message mentions the collapsing time line (and the end of the world) caused by arresting the wrong suspect (see Fig.~\ref{fig:backstory_outro}). We decided to use a suspect/culprit mechanic because of the interface it creates for the player to interact with the raw data: players must fact-check each suspect, therefore they have to collect and verify each fact to complete the game.

\begin{figure}
\begin{center}
\includegraphics[width=0.95\columnwidth,clip,trim={2cm 1.5cm 2cm 3cm}]{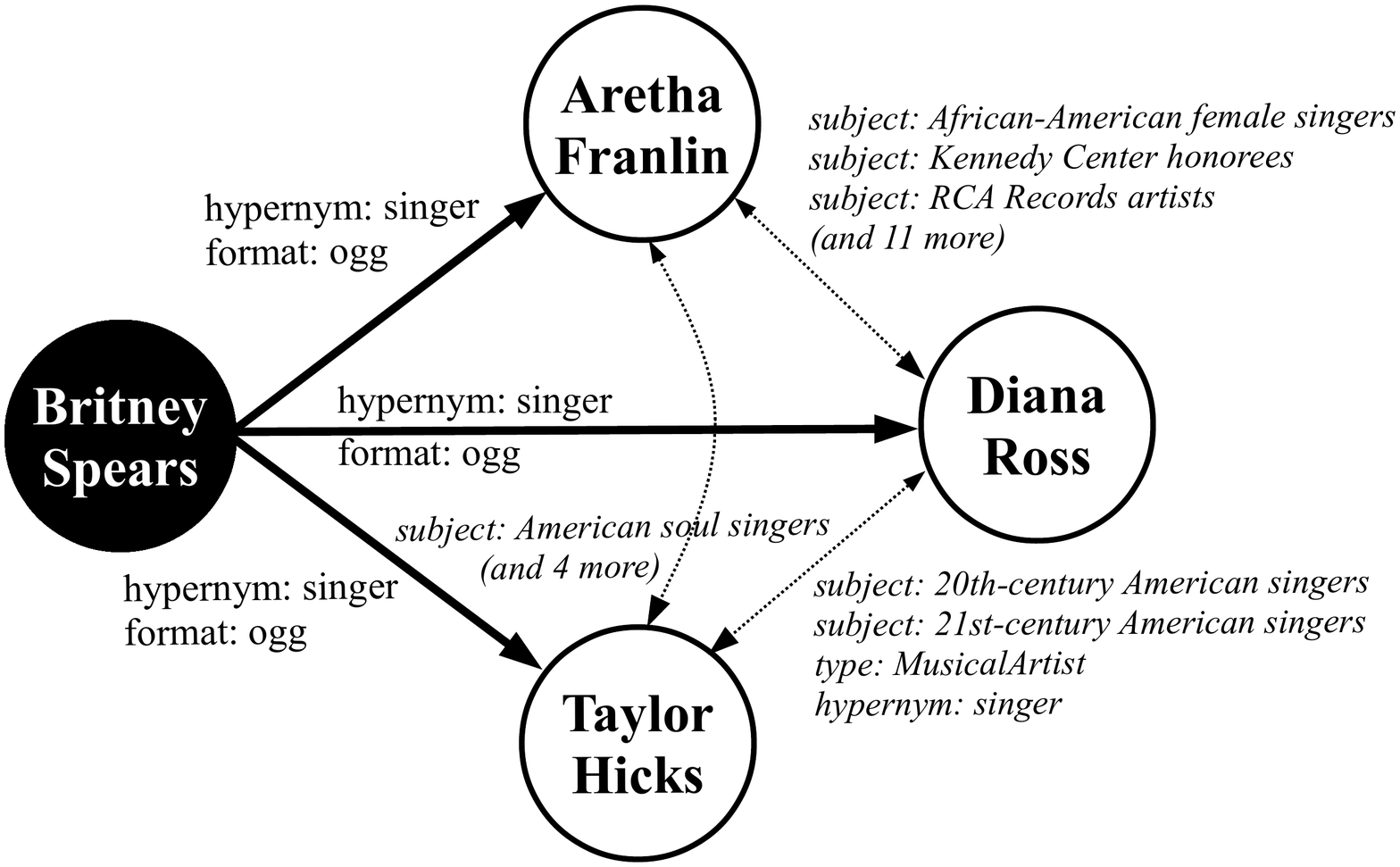}
\end{center}
\caption{Example for suspect selection and its fitness calculation, in the Britney Spears testbed game. The fitness attempts to maximize the common predicates and objects between the victim and the suspects (bold arrows) and between the suspects themselves (dotted arrows, italics). Links are shown in the format ``${\langle}$predicate${\rangle}$: ${\langle}$object${\rangle}$''.}
\label{fig:fitness-illustration}
\end{figure}

\subsection{Generating Adventures}\label{sec:generator}

The system generates an adventure seeded by a person's name. A character representing this person's Wikipedia article becomes the victim of the murder, around which the plot is built. Initially, the system finds a large pool of possible suspects, and narrows it down to a select group of suspects that will appear in the game. Then, it finds paths linking each suspect and the victim, which eventually will shape the game's plot. The next sections explain this process.

\subsubsection{Suspect Selection}\label{sec:generator_suspects}

\begin{table*}[tb]
\centering
\caption{Example dialog lines for different sentences. ``Type'' defines whether the dialogue is \emph{Essential}, \emph{Fact-giving}, or \emph{Flair}, and ``Speaker'' designates who utters the line in-game (Player or NPC). Words in brackets are instantiated in each game based on information for this NPC or the plot of the adventure.}
\label{table:grammar}
\small
\begin{tabular}{l|l|l|l}
\textbf{Sentence} & \textbf{Example Lines} & \textbf{Type} & \textbf{Speaker} \\ \hline \hline
Greeting-to-Player & ``Hello? Can I help you?'', ``Hi. Do you need something?'', ``Hi.'' & Essential & NPC \\ \hline
Central-Hub & \begin{tabular}[c]{@{}l@{}}``Greetings, I'm a private-eye. I'm sorry to disturb, could I ask you some questions?'',\\ ``Greetings, I'm a detective. Can I ask you some questions?''\end{tabular} & Essential & Player \\ \hline
Central-Hub Response & ``Okay. Ask away.'', ``Sure. I'll do my best.'', ``Of course. I can try.'' & Essential & NPC \\ \hline
Name-Query & \begin{tabular}[c]{@{}l@{}}``Please, tell me your name.'', ``Who are you?'', \\ ``Would you tell me your name?''\end{tabular} & Flair & Player \\ \hline
Name-Response & \begin{tabular}[c]{@{}l@{}}``Yes, I'm ${\langle}$\texttt{personName}${\rangle}$'', ``I am known as ${\langle}$\texttt{personName}${\rangle}$.'', \\ ``I am called ${\langle}$\texttt{personName}${\rangle}$.''\end{tabular} & Flair & NPC \\ \hline
Clue-Query & \begin{tabular}[c]{@{}l@{}}``Did the victim remark anything that you can remember?'',\\ ``Is there something you think I should know about?'',\\ ``Do you have something that might be useful to the investigation?''\end{tabular} & Essential & Player \\ \hline
Clue-Response-Building & \begin{tabular}[c]{@{}l@{}}``The victim often traveled to a place named ${\langle}$\texttt{building}${\rangle}$.'', \\ ``The victim often visited a place known as ${\langle}$\texttt{building}${\rangle}$.'', \\ ``I believe the victim talked about ${\langle}$\texttt{building}${\rangle}$.''\end{tabular} & Essential & NPC \\ \hline
Clue-Response-Concept & \begin{tabular}[c]{@{}l@{}}``If I remember correctly, the victim talked about a thing called ${\langle}$\texttt{thing}${\rangle}$.'', \\ ``If I recall correctly, the victim mentioned ${\langle}$\texttt{thing}${\rangle}$.'',\\ ``I think the victim mentioned a thing named ${\langle}$\texttt{thing}${\rangle}$.''\end{tabular} & Essential & NPC \\ \hline
Clue-Response-Place & \begin{tabular}[c]{@{}l@{}}``Last I thought, the victim went to ${\langle}$\texttt{place}${\rangle}$.'', \\ ``Last I heard, I heard the victim visited ${\langle}$\texttt{place}${\rangle}$.'', \\ ``Last I heard, I think the victim was seen in ${\langle}$\texttt{place}${\rangle}$.''\end{tabular} & Essential & NPC \\ \hline
Clue-Response-Person & \begin{tabular}[c]{@{}l@{}}``I believe the victim talked to ${\langle}$\texttt{personObject}${\rangle}$.'',\\ ``The victim was friends with ${\langle}$\texttt{personObject}${\rangle}$.'',\\ ``Last I heard, the victim was acquaintances with ${\langle}$\texttt{personObject}${\rangle}$. \\ Perhaps they know about something.''\end{tabular} & Essential & NPC \\ \hline
Suspect-Fact-Query & \begin{tabular}[c]{@{}l@{}}``Do you remember any information about one of the suspects?'',\\``Would you give me any information about one of the suspects?'',\\``I need some help. Would you tell me some information about one of the suspects?''\end{tabular} & Fact-giving & Player \\ \hline
Suspect-Fact-Response & \begin{tabular}[c]{@{}l@{}}``${\langle}$\texttt{suspectName}${\rangle}$ is an ${\langle}$\texttt{attribute}${\rangle}$.''\end{tabular} & Fact-giving & NPC \\ \hline
Residence-Query & ``Where do you currently reside? It might help with the investigation.'' & Flair & Player \\ \hline
Speculation-Response & ``I live in ${\langle}$\texttt{place}${\rangle}$.'', ``That's none of your business!'' & Flair & NPC \\ 
\end{tabular}
\end{table*}

At the beginning of the suspect selection process, the system performs a series of queries to find people related to the victim. These queries use DBPedia, a system that structures information from Wikipedia in the form of tuples~\cite{auer2007dbpedia}, and the people it finds become a pool of possible suspects. Suspects must have a direct link to the victim, sharing at least one characteristic value (object) with the them. Specifically, two DBPedia entries are linked if they have the same characteristic value assigned to the same predicate: for example, both Britney Spears and Diana Ross have a ``hypernym'' (predicate) with value of ``singer''. See Fig.~\ref{fig:fitness-illustration} for a sample of shared predicates and objects. The query for suspect pooling returns only DBPedia entities of type ``person''. Depending on the number of predicates and their values in the DBPedia entry of the victim, the pool of possible suspects can be very large. Indicatively, with Albert Einstein as the victim the suspect pool is 5,092 people; with Britney Spears as the victim it is 17,415.

The system uses an $\mu+\lambda$ evolution strategy with cascading elitism~\cite{togelius2007towards} over the suspect pool to select an optimal group of suspects. Each genome consists of a set of $N$ suspects (based on previous feedback on the game, this paper uses $N=3$ suspects per case). The system uses two fitness functions on a population of 50 individuals. The first fitness function is the number of direct links between the victim and all  suspects in the genome; the population is sorted based on this fitness and the worst half (lowest values) is removed. The remaining individuals are sorted based on the second fitness function, i.e. the number of direct links between suspects, and the worst half is again removed. This leaves only 25\% of the original population, and gives clear priority to the first objective (i.e. suspects connected to the victim) compared to the second objective (i.e. suspects connected to each other). The best 25\% of the original population is then copied and mutated to return the population to its original size. Mutation chooses one suspect in the genotype, and replaces it with a random entry from the suspect pool. The process ends after 500 generations, and the trio of suspects with the highest fitness is chosen as suspects for the game. Out of these, one is marked at random as the culprit.

Fig. \ref{fig:fitness-illustration} illustrates how each fitness is calculated: the score of the first fitness is 6 (2 common values between Britney Spears and each of the 3 suspects) and the score for the second fitness is 23 (14, 4 and 5 common values between the different suspects). Our assumption is that these fitnesses will result in a mystery where suspects have more common traits with the victim and between them. We experimented with a different function for the second sorting process, rewarding suspects with different characteristics, but the resulting sets were deemed too random and unintuitive. 


\subsubsection{Path Generation}\label{sec:generator_path}

The generator creates each game element, be it an NPC, item or location, from articles linking the victim to each suspect. For each suspect, a series of search queries is made to DBpedia in order to find paths of articles that link the victim to each suspect. Due to computational limitations in querying the DBpedia endpoint, we limited the maximum path length to 4 edges (i.e.~5 articles) per path. Each suspect can have multiple paths linking it to the victim: these paths are assessed based on the types of articles in one path compared to types found in all paths for the same victim-suspect pair. All paths are sorted based on their diversity and length, favoring those that are longer \textit{and} contain articles of different types (e.g. places, people, etc). For example, a path with 4 articles is preferred to one with 2; and a path with an article about a person and one about a place is preferred to one about two places. After sorting, the best path is selected for that victim and suspect. This approach presents small but important deviations from the previous versions of Data Adventures. The old systems repeat the path generation process once, essentially finding minor paths between articles along the main (major) path that links victim and suspect. This results in paths that are far longer, increasing gameplay time. 
Indicatively, for 3 suspects and a maximum path length of 4, DATA Agent would have $15$ objects if the system always selects a path of maximum length, while the WikiMystery generator \cite{barros2018killed} which precedes it would have $60$. While one may argue that longer games may be more interesting, playtesting showed us that when the player's interaction options are not varied enough, the game becomes repetitive and boring.
%

Once a best path is found for every suspect, the system generates in-game objects for each node in the paths: articles about locations are transformed into in-game cities and buildings; those about people become NPCs, and anything else is made into a game item, such as a book or a letter. Items and NPCs are placed in buildings within a city and are given descriptions, images, names, dialog lines as applicable. Items' descriptions and NPCs' dialog give clues that can reveal new buildings in the current city or a previously seen (or unseen) city, until finally the player can reach a suspect. When necessary, new NPCs or items are generated to fill gaps in the path (e.g. when an article is about a city, an NPC is created with a random name and minimal dialog and placed in a building within that city). Finally, lock and key puzzles are added to the game's paths using a variation of a breadth first algorithm that guarantees that each building can be unlocked with keys found before it. 

\subsubsection{Dialog Generation}\label{sec:generator_dialog}

DATA Agent uses the Tracery grammar system~\cite{compton2014tracery} to create dialog. There are three types of dialog, each requiring its own grammar: \emph{essential} dialog, \emph{fact-giving} dialog or \emph{flair} dialog. Essential dialog gives clues about game objects, progressing the story and making the locations containing them accessible. Fact-giving dialog reveals information about suspects which is important when the player is interrogating a suspect to fact-check their responses. Flair dialog gives information about the current NPC that the player is talking to, such as their birth date or what they are known for, and it is not essential to progress in the game. Not every NPC has fact-giving or flair dialog, and suspects do not have dialog at all: their conversation triggers an interrogation, following a different mechanic than previous character interactions. Table \ref{table:grammar} defines several sentences used in the game, possible example lines that could be generated for this sentence, the sentence's dialog type, and the speaker of the sentence. 


\begin{figure*}[t]
\centering
\subfloat[Singers at the house of the deceased, in McComb, Mississippi (the birthplace of Britney Spears)]{\includegraphics[width=0.32\textwidth]{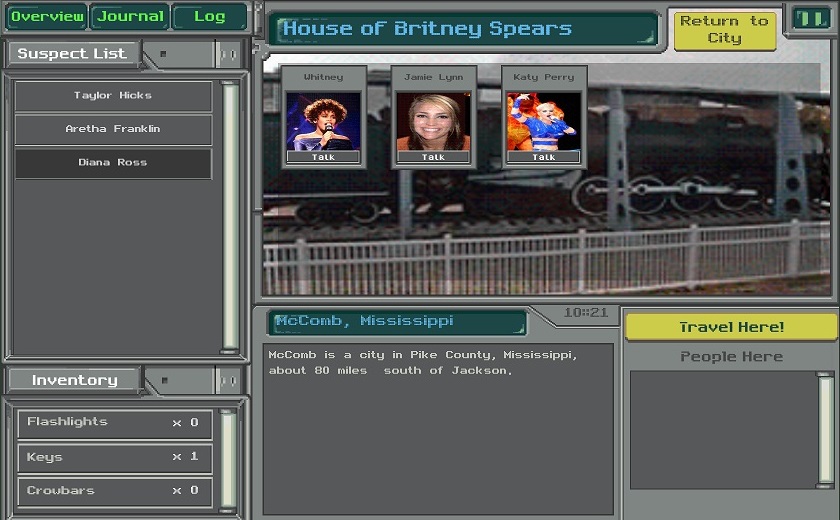}\label{fig:s1}}\quad
\subfloat[A suspect makes a claim which can be falsified]{\includegraphics[width=0.32\textwidth]{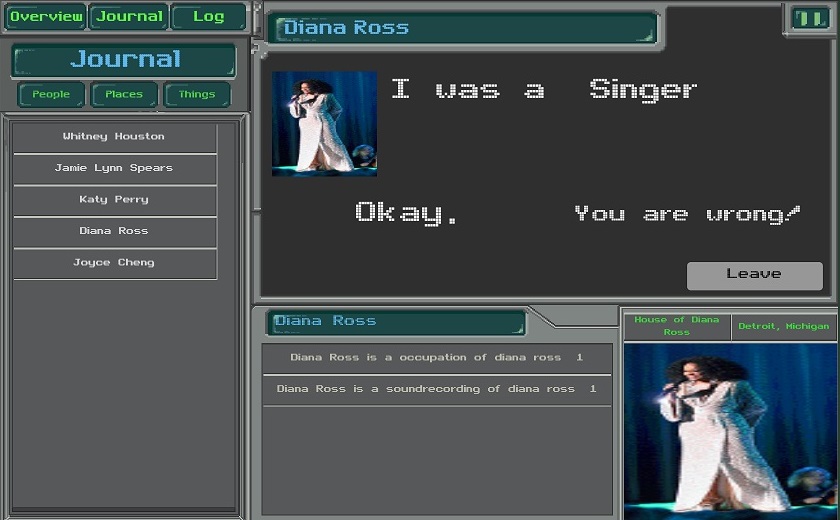}\label{fig:s2}}\quad
\subfloat[An item consisting of a very sparse list]{\includegraphics[width=0.32\textwidth]{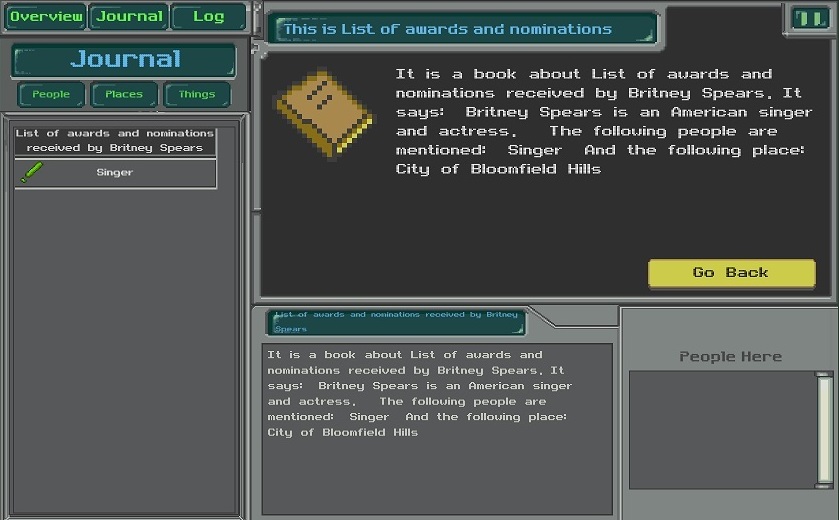}\label{fig:s3}}
\caption{Screenshots of the playtested Britney Spears game.}
\label{fig:britney-spears-game}
\end{figure*}

\section{User Study}\label{sec:study}

In order to test the games generated by the DATA Agent system, a user study was conducted with 30 participants playing two DATA Agent games. The study was largely exploratory, and its goals were to assess whether the design choices in both the user interface and the generator made the game easier to use (in terms of usability) and intuitive to solve (in terms of the difficulty in solving the mystery and the extent of interactions with the generated game content). The protocol followed, the games tested, the participants' demographics, and an analysis of their gameplay and feedback is provided in the following sections. Any statistical significance analysis reported is made with $\alpha=0.05$ threshold.

\begin{table}[tb]
\centering
\caption{Average interaction ratios for all playthroughs, along with their 95\% confidence intervals. The normalized (nrm.) values are divided by the number of respective game objects in each game.}
\label{table:gameplay-metrics}
\small
\begin{tabular}{l|ccccc}
\textbf{Values} & \textbf{City} & \textbf{Building} & \textbf{Item} & \textbf{People} & \textbf{Dialog} \\
\hline \hline
\multicolumn{6}{l}{Albert Einstein game}\\
\hline
Raw & 8.4$\pm$2.4& 15.5$\pm$2.8& 9.7$\pm$0.8& 5.8$\pm$0.9& 43$\pm$5.7 \\ 
Nrm. & 2.8$\pm$0.8& 1.6$\pm$0.3& 1.1$\pm$0.1& 0.8$\pm$0.1& 1$\pm$0.1 \\
\hline \hline
\multicolumn{6}{l}{Britney Spears game}\\
\hline
Raw &10.6$\pm$2& 16.1$\pm$2.6& 7.3$\pm$0.4& 9$\pm$1.1& 66.7$\pm$8.8 \\
Nrm. & 2.1$\pm$0.4& 1.6$\pm$0.3& 1$\pm$0.1& 0.9$\pm$0.1& 1.3$\pm$0.2 \\
\end{tabular}
\end{table}

\subsection{Experiment Design}\label{sec:study_protocol}

Subjects were asked to individually playtest two games generated by DATA Agent and give their opinions about their experience. Participants were found through use of New York University e-mail lists and social media, through which we asked for volunteers. All participants were brought to a quiet classroom setting. Before the test, participants were informed about open data, data games, and the motivations behind the research. It was stressed that everything in the DATA Agent games had been generated from open data using artificial intelligence and had not been hand-authored.

Subjects first filled out a pre-game questionnaire to obtain demographic information (gender, age), play habits (frequency and preference for games) and experience with adventure games (preference and three example adventure games). 
The subjects were then asked to play two DATA Agent mysteries: \textbf{The Case of Albert Einstein}, and \textbf{The Case of Britney Spears}. These two cases were selected for their difference in terms of gender, type of celebrity, and contemporaneity. The order of games was randomly selected for each participant.
During play, participants were asked to freely voice their opinions about the game and the data, as all playthroughs were audio recorded. Researchers were available in case the player needed help or had questions about gameplay. All in-game player actions were logged, and specifically game events (e.g. cities visited) were logged and analyzed in Section \ref{sec:analysis}.  
After the participants completed both mysteries (regardless of whether they won or not), participants were asked to fill out a post-game survey. The survey contained 11 questions regarding the gameplay and the broader appeal: answers were on a 5-scale Likert scale from ``strongly disagree'' to ``strongly agree''. Additionally, the survey included 4 questions regarding broader suggestions and reflections: answers to the latter were as free-form text.
After the experiment was over, participants were compensated with a \$10 Amazon Gift Card.

\subsection{Tested Games}\label{sec:study_games}
In \textbf{The Case of Albert Einstein} (see Fig.~\ref{fig:interface}), players set out to solve the murder of Albert Einstein, for which the suspects are Walther Bothe, Lise Meitner, and Max von Laue; Lise Meitner is the (doppelganger) culprit. 
This game involves regular NPCs such as Langston Hughes, Franklin D. Roosevelt, and Harry Emerson Fosdick. The player can travel between three cities: Switzerland, The German Empire, and Germany. In total, there are 10~buildings (e.g. ``House of Lise Meitner'' in Fig.~\ref{fig:d5}, ``Library'' in Fig.~\ref{fig:d2}), 9~items (e.g. ``a photograph about The Evolution of Physics: The Growth of Ideas From Early Concepts to Relativity and Quanta'' in Fig.~\ref{fig:d3}), 7~people, and 41~dialog nodes to explore.

In \textbf{The Case of Britney Spears} (see Fig.~\ref{fig:britney-spears-game}), players set out to solve the murder of Britney Spears, for which the suspects are Aretha Franklin, Diana Ross, and Taylor Hicks; Diana Ross is the (doppelganger) culprit.
This game involves regular NPCs such as Whitney Houston, Jamie Lynn Spears, and Katy Perry. The player can travel between five cities: ``McComb, Mississippi''; ``New York City''; ``Birmingham, Alabama''; ``City of Bloomfield Hills''; and ``Detroit, Michigan''. In total, there are 10~buildings, 7~items (e.g. ``a book about List of awards and nominations received by Britney Spears'' in Fig.~\ref{fig:s3}), 10~people, and 52~dialog nodes to explore.

\subsection{User Data}\label{sec:study_users}
\begin{table}[tb]
\centering
\caption{Weighted average (1 for strongest disagreement, 5 for strongest agreement) of user's responses to specific post-game questions and its 95\% confidence interval. Also shown are the number of positive (agree and strongly agree) versus negative (disagree and strongly disagree) responses.}
\label{table:usability-metrics}
\small
\begin{tabular}{p{5.2cm}|c|c@{}c}
\textbf{Question} & \textbf{Average} & \textbf{Pos}./& \textbf{Neg}.\\
\hline \hline
DATA Agent is an adventure game. & 3.77$\pm$0.35 & 21 & 3 \\ \hline	 DATA Agent is fun, relative to other adventure/role-playing games I have played. & 3.4$\pm$0.26 & 15 & 2 \\ \hline	
DATA Agent is challenging, relative to other adventure/role-playing games I have played. & 2.37$\pm$0.35 & 4 & 17 \\ \hline	
It is easy to understand how to play DATA Agent. & 3.87$\pm$0.35 & 22 & 4 \\ \hline	
I learned something new playing DATA Agent. & 3.53$\pm$0.34 & 20 & 5 \\ \hline	
I would get DATA Agent if it were a fully developed as a mobile or computer game. & 3.4$\pm$0.39 & 16 & 6 \\ \hline	
I would recommend DATA Agent to a friend if it were fully developed as a mobile or computer game. & 3.6$\pm$0.32 & 19 & 3 \\ \hline	
I would like to play DATA Agent again with the same mystery. & 2.3$\pm$0.39 & 5 & 20 \\ \hline	
I would like to play DATA Agent again with a different mystery. & 3.83$\pm$0.38 & 24 & 5 \\ \hline	
Finding the culprit in DATA Agent was straightforward and made sense to me. & 3.9$\pm$0.37 & 21 & 4 \\ \hline	
Finding the culprit in DATA Agent requires mental effort. & 2.6$\pm$0.36 & 7 & 15 \\ 
\end{tabular}
\end{table}
In total, 30 adults (22 identifying as male, 7 as female, 1 as other) participated in the user study. The median participant age was 27.5, but most participants (53\%) were between 19-25 years old. An overwhelming majority (93\%) of playtesters reported that they enjoyed playing games, and a slightly smaller majority (83\%) reported enjoying playing adventure games specifically. Two thirds of playtesters claimed to play video games at least as often as 2-3 times a week, while only 13\% of players played less often than once a week.  


\subsection{Gameplay Data}\label{sec:analysis}

Overall, 23 participants managed to find the culprit in both games while 28 participants found the culprit in at least one game. There was no clear difference between the number of people who solved the Albert Einstein case (25 players) over the Britney Spears case (26 players). How users interacted with different game entities in each game is shown in Table~\ref{table:gameplay-metrics}. 
Obviously, players interacted more with people and cities in the Britney Spears game where there were more people and cities. Normalizing interactions based on the number of game objects of that type shows that the differences between the two games were not as pronounced. However, it is obvious that the Britney Spears game required more player attention in general; based on general feedback, that game required a longer playtime as well. 
In general, it was surprising to us that the average interactions with NPCs is smaller than the number of NPCs of each game, meaning that some NPCs were not interacted with at all. Indeed, the NPC interactions were fewer than the NPCs in 73\% of Einstein games and in 63\% of Spears games. It is not surprising that players visited cities more than once, however, as each new clue often revealed a new building in existing cities (or a new NPC in an existing building) and required players to travel back to previously visited cities.
It should be noted that players had very different interaction patterns: for instance, in the Albert Einstein game players interacted between 3 and 15 times with NPCs. It should also be noted that the two players who had only 3 NPC interactions managed to win the game. In general a correlation analysis between the raw or normalized gameplay values of Table~\ref{table:gameplay-metrics} and whether the game was won showed no significant correlations.



\subsection{Usability Data}\label{sec:study_usability}

For the sake of brevity, we discuss only the responses of users which were provided in a Likert scale. The results of these responses are summarized in Table~\ref{table:usability-metrics}. For the sake of analysis, we consider the average rating on the Likert scale to be misleading, and instead we use the number of positive responses (``agree'' or ``strongly agree'') versus the number of negative responses (``disagree'' or ``strongly disagree'') as a more granular measure which shows the inter-rater agreement. Overall, the number of ``neutral'' answers were low, with a notable exception regarding whether ``DATA Agent is fun\ldots'' (with 43\% of answers being ``neutral''). Similarly, participants tended to agree in terms of either positive or negative responses. Considering positive responses as successes (or negative responses, for the reverse), the one-tailed binomial test (with $\alpha=5\%$) is passed for the following assertions:
%
\begin{enumerate}
\item DATA Agent \textbf{is} an adventure game.   


\item DATA Agent \textbf{is not} challenging, relative to other adventure/role-playing games I have played.  

\item It \textbf{is} easy to understand how to play DATA Agent.     

\item I \textbf{did} learn something new playing DATA Agent.

\item I \textbf{would} recommend DATA Agent to a friend if it were fully developed as a mobile or computer game.   

\item I \textbf{would not} like to play DATA Agent again with the same mystery.

\item I \textbf{would} like to play DATA Agent again with a different mystery.  

\item Finding the culprit in DATA Agent \textbf{was} straightforward and made sense to me.   
\end{enumerate}
%

Similarly to gameplay data, we investigated whether the user's responses correlate with their winning streak. The assertions ``I learned something new\ldots'' and ``finding the culprit [\ldots] was straightforward\ldots'' had a significant positive correlation with the number of games won by that player ($r=0.36$ and $r=0.46$ respectively). Moreover, significant positive correlations were found between the usability assertion ``It is easy to understand how to play DATA Agent.'' and agreement regarding ``In general, I like to play games.'' ($r=0.44$) and ``I like to play adventure games'' ($r=0.53$). These correlations should not be surprising, as the interface design of DATA Agent was inspired by classic point-and-click adventure games; users familiar with games and especially with adventure games would be more likely to find this game easier to play.


\section{Discussion}\label{sec:discussion}

The conclusions drawn from a user study with 30 participants are generally positive. Most participants agree that DATA Agent is easy to understand, and solving the mystery was straightforward. The major overhaul of the user interface (compared to previous titles in the Data Adventures series) seems to now be intuitive to use, especially by users more familiar with games in general and adventure games in particular. While the majority of players would not try the game again with the same mystery, they would like to try the game with another mystery. This is hardly surprising, as adventure games in general are rarely replayable. With the small set of interactions (few suspects, people, and cities), participants in this version of DATA Agent likely had explored all possible aspects of each mystery on their first try. It can be assumed, on the other hand, that a generator such as the one in DATA Agent, which can create infinite mysteries, can keep players engaged for a long period of time. We can assume that a large part of this appeal is not the fun gameplay itself (which we discuss later), but the players' natural curiosity regarding which associations the generator will make and how those will be presented in terms of visuals and dialog.

On the other hand, the participants' responses highlighted that the game now is perhaps too easy. Most participants agree that the game is not challenging. We could also interpret the large number of neutral responses in terms of how fun the game is, or the agreement that finding the culprit is straightforward as further proof that the straightforward solutions offered by the game make it feel ``bland''. To a degree, many of the design decisions taken when generating the two testbed games were explicitly to reduce the length and complexity of gameplay. The smaller number of suspects (three versus five in earlier efforts \cite{barros2018killed}) and the shorter paths (without minor paths as found in \cite{barros2015data,barros2018killed}) all contribute to a shorter playtime and a more direct way to reach each suspect. This reduces the time investment of interested participants and allows us to receive more feedback from more users faster. Additionally, it highlights important information, that would otherwise be lost amidst a large amount of content: WikiMystery generates 135 game objects on average per adventure \cite{barros2018killed} (NPCs, locations and items) while DATA Agent cuts that down to 30 game objects. However, it is not easy to estimate how to increase the challenge of the game without making it unplayable or banal. Re-introducing more suspects would increase the number of people, buildings, cities and items that the player has to interact with (prolonging playtime) and may increase the challenge as players will need to interrogate and find facts for more suspects. On the other hand, this could result in more ``trivial'' interactions that feel inconsequential, and could clutter the interface (e.g. by having 30 or more people in the Journal Tab with little ability for the player to understand which one has been recently added (or may hide more information). Towards increasing the ``mystery'' aspect of the game, the vital facts for interrogating suspects could be hidden away in more locations (possibly locked ones) instead of being offered by NPCs at the house of the deceased. Moreover, having multiple ways to reach the same fact or suspect could make the choices of the players more meaningful; for instance, the game could have one key that can open two locks (i.e.~the player will only have access to one but not both of those buildings) so the game can be solved with the clues hidden in either but not both of those locations. Other additions, such as a broader set of dialog options for flair, are likely to enhance the curiosity and learning potential that motivates and benefits the player respectively.

In terms of broader directions for taking DATA Agent forward, the game design and algorithmic implementation should capitalize on the strengths noted by participants in the user study and minimize the negative factors such as the perceived lack of challenge. As noted, more extensive dialog could lead to more interesting interactions: such dialog generation could take further advantage of open data (allowing users to ask more from each NPC beyond their name and the subject they are known for), or follow authored templates to introduce more murder mystery tropes such as other NPCs offering potential motives or alibis for the suspects based on their relationship with the suspect (which can be calculated from open data, or not). Facts about the suspects could be chosen more wisely, since certain facts are nonsensical (e.g. in Fig.~\ref{fig:s2} one of the facts for Diana Ross is ``Diana Ross is a soundrecording of dianaross~1'') or trivial to ascertain the veracity of (e.g. ``I was a singer'' in Fig.~\ref{fig:s2}). This finding is corroborated by the user survey, as players managed to solve each game despite talking to a small subset of all NPCs. Finally, the cities used in the game could be improved, both in terms of name (e.g. players can travel to ``The German Empire'' in the Einstein game, which is hardly a city) and in terms of the city map used to display them (e.g. in Fig.~\ref{fig:d2}). Information from OSM could be used directly to place actual buildings in their correct location on the map, rather than at random as it is currently. The buildings from OSM could show a generic background (e.g. a generic ``Supermarket'' background if the OSM location says that this location is a shop of any type) or could be specific to the location based on an image search with that specific location keyword (e.g. an image of the ``Coliseum'' at its exact location in Rome).

Since the previous generation of the Data Adventure series \cite{barros2018killed} there have been substantial improvements with how the data is presented to the player. However, generating games from open data is still heavily dependent on the uncontrollable nature of said data. Games often include facts that are nonsensical, as pointed out above (``Diana Ross is a soundrecording of dianaross~1''). Sometimes the image for a character is not correct, and several times throughout the user study, players commented on this fact. However, we find it nearly impossible to ever fully control or constrain these occurrences, since doing so would involve handpicking data, defeating the purpose of open data games. In the long run, an important question is to what extent this type of data-driven game generators can ever be trusted to be fully autonomous, given the bizarre and sometimes even offensive situations and characters that can occasionally emerge even from correct underlying data.


\section{Conclusion}\label{sec:conclusion}

This paper presented the DATA Agent system, which generates adventures based on Wikipedia. These adventures task the player with solving the fictional murder of a real-world person. The system utilizes linked data to discover appropriate suspects for the murder within Wikipedia, and places these real-world people, locations, and encyclopedic subjects in a fictional setting where time and space are relative. DATA Agent is the latest installment in the Data Adventures series, and comes with an extensive overhaul of the user interface and the exposition of the game's backstory. Coupled with refinements in the dialog and interrogation system and a simplification of the generative algorithms for creating a mystery from open data, the resulting games are shown to be easy to grasp, straightforward to solve, and generally fun to play. Based on a user study with 30 participants, two games generated for DATA Agent were deemed fairly easy to solve, based on a high completion ratio as well as user feedback in the end of the experiment. Future work should aim to introduce more challenges in the gameplay, either by making the volume of interlinked content (and suspects) larger or by obfuscating the existing solutions in more complex dialog trees or puzzles (spatial or combinatorial).

The system improves on several facets of previous Data Adventures installments, but still comes short on some aspects. The new backstory helps flesh out a narrative and give the player a more engaging and consistent scenario, but much work could be done to improve the experience. On a narrative level, the system could integrate subplots into the main story, such as love affairs or intrigues among NPCs. Allowing NPCs to have personal goals, personalities, feelings or even freedom of movement would provide a deeper sense of immersion for players. Additionally, there is room for puzzle generation that takes into account not only the raw source data, but also the general narrative choices made by the generator. For example, if a generated subplot leans towards the rivalry between two NPCs based on real musicians (e.g. Mozart and Saliery), the system could generate a puzzle involving stealing a composition from one and giving it to the other. Finally, there is the issue of the absurdity present in open data-driven game generation, and the dilemma of further limiting absurdity without constraining the expressivity of the system.

\section*{Acknowledgments}
Gabriella Barros acknowledges financial support from CAPES and Science Without Borders program, BEX 1372713-3, as well as an NYU Tandon School of Engineering Fellowship. Antonios Liapis has received funding from the European Union's Horizon 2020 research and innovation programme under grant agreement No 693150. Michael Green acknowledges financial support from the GAANN program.

\bibliographystyle{acm}
\bibliography{data_agent}

\end{document}